\documentclass[10pt,twocolumn,letterpaper]{article}

\usepackage[table]{xcolor}
\definecolor{gray!20}{gray}{0.8} 
\definecolor{green!50!black}{RGB}{0,128,0} 
\definecolor{red}{RGB}{255,0,0} 

\usepackage{cvpr}
\usepackage{times}
\usepackage{epsfig}
\usepackage{graphicx}
\usepackage{amsmath}
\usepackage{amssymb}
\usepackage{microtype}
\usepackage{booktabs} 
\usepackage{svg}
\usepackage{pifont}

\newcommand{\xmark}{\ding{55}} 
\newcommand{\gcheck}{\textcolor{green!50!black}{\checkmark}}
\newcommand{\rcross}{\textcolor{red}{\xmark}}

\usepackage{mathtools}
\usepackage{amsthm}
\usepackage{algorithm}
\usepackage{algpseudocode}
\usepackage{multirow}
\usepackage{tikz}
\usepackage{pgfplots}
\usepackage[normalem]{ulem}

\theoremstyle{plain}

\theoremstyle{definition}

\theoremstyle{remark}

\usepackage[breaklinks=true,bookmarks=false]{hyperref}

\cvprfinalcopy 

\def\cvprPaperID{****} 

\begin{document}

\title{SGD-Mix: Enhancing Domain-Specific Image Classification with Label-Preserving Data Augmentation}

\author{
Yixuan Dong\textsuperscript{1}\thanks{Equal contribution.} \quad
Fang-Yi Su\textsuperscript{1,2}\footnotemark[1] \quad
Jung-Hsien Chiang\textsuperscript{1} \\
\textsuperscript{1}Department of Computer Science and Information Engineering,\\
National Cheng Kung University, Tainan City 701, Taiwan, R.O.C.\\
\textsuperscript{2}Department of Biomedical Informatics,\\
Harvard Medical School, Harvard University, Boston, MA, USA\\
{\tt\small radondong@iir.csie.ncku.edu.tw, fang-yi\_su@hms.harvard.edu}
}

\maketitle

\begin{abstract}
Data augmentation for domain-specific image classification tasks often struggles to simultaneously address diversity, faithfulness, and label clarity of generated data, leading to suboptimal performance in downstream tasks. While existing generative diffusion model-based methods aim to enhance augmentation, they fail to cohesively tackle these three critical aspects and often overlook intrinsic challenges of diffusion models, such as sensitivity to model characteristics and stochasticity under strong transformations. In this paper, we propose a novel framework that explicitly integrates diversity, faithfulness, and label clarity into the augmentation process. Our approach employs saliency-guided mixing and a fine-tuned diffusion model to preserve foreground semantics, enrich background diversity, and ensure label consistency, while mitigating diffusion model limitations. Extensive experiments across fine-grained, long-tail, few-shot, and background robustness tasks demonstrate our method’s superior performance over state-of-the-art approaches.
\end{abstract}

\section{Introduction}
\label{Introduction}

In the early stages of data augmentation research for domain-specific image classification tasks~\cite{mumuni2022data,litjens2017survey,shorten2019survey}, non-generative mixup-based methods (e.g., Mixup~\cite{zhang2018mixup} and CutMix~\cite{yun2019cutmix}) typically linearly blend two images and assign labels to the generated data based on their mixing ratios. However, due to constraints imposed by the linear sample space and issues such as unnatural overlaps in salient regions and unclear boundaries, the generated data often fail to ensure sufficient diversity and faithfulness. Moreover, determining labels solely based on mixing ratios can lead to label-semantic mismatches~\cite{araslanov2021self,qin2024sumix}, ultimately hindering performance improvements in downstream tasks.

\begin{figure}[!t]
    \centering
    \includegraphics[width=\columnwidth]{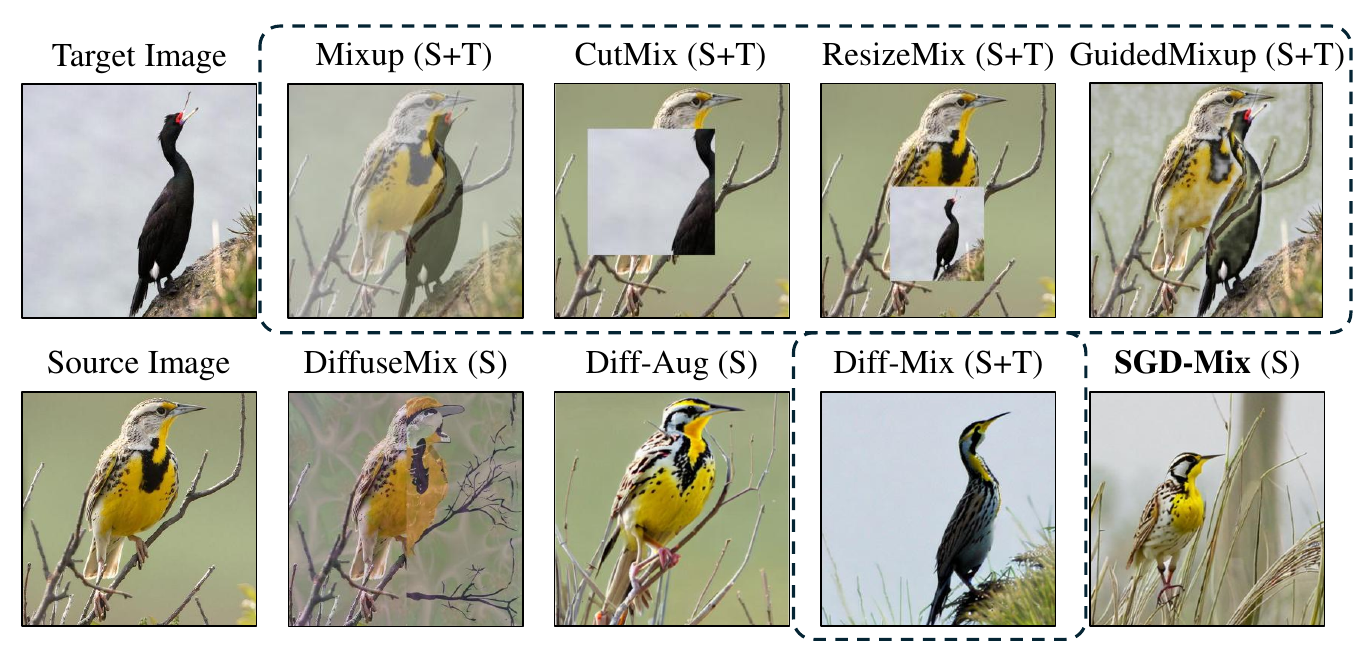}

    \caption{Top row: Non-generative mixup-based methods. Bottom row: Generative diffusion-based methods. Inside the dashed box: Methods that mix two different training images and their corresponding labels. Outside the dashed box: Label-preserving methods without label mixing. Note that the translation strength for the bottom row in the figure is consistently set to 0.7.}
    \label{fig:mixup_comparison}

\end{figure}

Subsequent non-generative improvements, such as GuidedMixup~\cite{kang2023guidedmixup}, which employs saliency-guided pixel-level fusion, and SUMix~\cite{qin2024sumix}, which integrates semantic distance metrics, have helped alleviate issues related to unnatural overlaps and label-semantic mismatches, thereby improving the quality of the generated data as well as the reliability of label assignments. A growing body of evidence identifies \textbf{diversity}, \textbf{faithfulness}, and \textbf{label clarity} as key factors in the quality and effectiveness of generated data. However, these methods still struggle to enhance these three factors simultaneously.  

With the advent of generative models~\cite{kingma2013auto,goodfellow2014generative,ho2020denoising}, particularly diffusion models~\cite{ho2020denoising,nichol2021improved,songdenoising,songscore} capable of nonlinear and high-fidelity synthesis~\cite{rombach2022high,saharia2022photorealistic,dhariwal2021diffusion,nichol2022glide}, a new direction for data augmentation in domain-specific image classification tasks~\cite{azizisynthetic} has emerged. Generated data produced by diffusion models exhibits greater diversity and faithfulness, surpassing previous non-generative mixup-based and GAN-based methods~\cite{zhang2018mixup,yun2019cutmix,kang2023guidedmixup,qin2020resizemix,kim2020puzzle,antoniou2018data}. Contemporary state-of-the-art (SOTA) diffusion-based methods such as Diff-Mix~\cite{wang2024enhance} and DiffuseMix~\cite{islam2024diffusemix} have shown promising results.

However, deeper analysis reveals potential limitations: like most existing approaches, these methods, in their design, lack a comprehensive consideration of diversity, faithfulness, and label clarity in generated data. Furthermore, they overlook inherent constraints imposed by the characteristics of diffusion models, which may hinder their ability to fully address these three critical aspects of generated data. We further analyze these methods and their limitations in Section~\ref{motivation} and \textcolor{magenta}{Supplementary Materials 1 \& 2}.

Overall, whether considering traditional non-generative or emerging generative methods, most solutions focus on only one or two key factors, lacking a holistic approach that concurrently addresses the diversity, faithfulness, and label clarity of generated data. For a summary of how existing approaches balance these attributes, please refer to Table~\ref{table:comparison}.

In response, we propose \textbf{Saliency-Guided Diffusion Mix (SGD-Mix)}, a novel framework that simultaneously ensures the diversity, faithfulness, and label clarity of generated data. As illustrated in Figure~\ref{fig:mixup_comparison}, SGD-Mix achieves label-preserving, high-quality data augmentation by retaining the semantic foreground of the source image while incorporating a diverse, contextually consistent background from another image, thus avoiding label ambiguity. This is accomplished by leveraging saliency maps~\cite{itti2002model} to align and mask the source and target images, strategically preserving the foreground of the source image while using the target image as the background. The resulting masked image is then refined by a domain-specific fine-tuned diffusion model. By adjusting the translation strength, SGD-Mix flexibly balances the diversity and faithfulness of the generated images, consistently ensuring that the labels remain unambiguous.

\begin{table}[!t]
\centering
\footnotesize
\resizebox{\columnwidth}{!}{
\begin{tabular}{lccc}
\toprule
\textbf{Method} & \textbf{Diversity} & \textbf{Faithfulness} & \textbf{Label Clarity} \\
\midrule
Mixup~\cite{zhang2018mixup}       & \rcross & \rcross & \rcross \\
GuidedMixup~\cite{kang2023guidedmixup} & \rcross & \gcheck & \rcross \\
DiffuseMix~\cite{islam2024diffusemix}  & \rcross & \gcheck & \rcross \\
Diff-Aug~\cite{wang2024enhance}    & \rcross & \gcheck & \gcheck \\
Diff-Mix~\cite{wang2024enhance}    & \gcheck & \gcheck & \rcross \\
\rowcolor{gray!20}\textbf{SGD-Mix (Ours)} & \gcheck & \gcheck & \gcheck \\
\bottomrule
\end{tabular}%
}
\caption{Comparison of representative data augmentation methods with respect to Diversity, Faithfulness, and Label Clarity. A green check (\gcheck) indicates that the method guarantees the corresponding factor, while a red cross (\rcross) indicates it does not.}
\label{table:comparison}
\end{table}

In summary, our key contributions are:
\begin{itemize}
    \item We first explicitly identify and emphasize the importance of simultaneously considering the \textbf{diversity}, \textbf{faithfulness}, and \textbf{label clarity} of generated data for effective data augmentation in domain-specific image classification tasks, rather than focusing on only one or two of these aspects.
    \item We propose a simple yet effective data augmentation framework that simultaneously meets all these three criteria, demonstrating superior performance across various domain-specific image classification tasks.
    \item We conduct comprehensive experiments and analyses, including detailed comparisons with previous methods and thorough ablation studies, to validate the effectiveness and robustness of our proposed approach.
\end{itemize}

\section{Related Work}
\label{related_work}

\textbf{Saliency Detection and Thresholding Methods.}  
Saliency maps play a crucial role in computer vision tasks by highlighting the most informative regions in an image. Early approaches~\cite{itti2002model} modeled saliency based on low-level features such as color, intensity, and orientation. Subsequent advancements incorporated spectral residual analysis~\cite{hou2007saliency} and deep learning-based gradient methods~\cite{simonyan2014deep,zhao2015saliency,selvaraju2017grad,zhou2016learning}, improving the identification of class-discriminative regions. In many applications, thresholding techniques such as Otsu’s method~\cite{otsu1975threshold} are used to convert saliency maps into binary masks for segmentation, object detection, and data augmentation.  

\textbf{Mix-Based Data Augmentation.}  
Mix-based augmentation techniques aim to improve model robustness by generating new training samples through the combination of existing ones. Mixup~\cite{zhang2018mixup} introduced a linear interpolation strategy to blend images and labels, while CutMix~\cite{yun2019cutmix} replaced rectangular regions between images to preserve more semantic structure. ResizeMix~\cite{qin2020resizemix} further refined this approach by incorporating scale transformations. More recently, methods like PuzzleMix~\cite{kim2020puzzle} and GuidedMixup~\cite{kang2023guidedmixup} have leveraged saliency information to enhance semantic consistency in mixed samples. These methods have been widely used in domain-specific image classification tasks and continue to evolve with more sophisticated strategies for region selection and blending.  

\textbf{Generative Model-Based Data Augmentation.}  
Generative models have provided a powerful alternative to mix-based augmentation by directly synthesizing new training samples. GANs~\cite{goodfellow2014generative,antoniou2018data} were among the first models used for this purpose, generating diverse data distributions to improve model generalization~\cite{antoniou2018augmenting}. More recently, diffusion models~\cite{ho2020denoising,nichol2021improved,songdenoising,songscore} have emerged as a state-of-the-art approach, offering high-fidelity image generation with strong controllability. Several data augmentation methods, such as Diff-Mix~\cite{wang2024enhance} and DiffuseMix~\cite{islam2024diffusemix}, have utilized diffusion models to generate augmented samples through noise scheduling and prompt-based conditioning~\cite{chen2023importance,dhariwal2021diffusion}. These approaches demonstrate the potential of generative models in enhancing domain-specific classification tasks.  

\section{Preliminary}
\label{preliminary}

\subsection{Image-to-image Translation Through Diffusion Models}
Let $\mathbf{x}_0^{\mathrm{ref}}$ be a reference image. We define a forward process that 
injects noise up to time step $\lfloor sT \rfloor$ (where $0 \le s \le 1$ and $T$ is the total number of timesteps), generating
\begin{equation}
    \mathbf{x}_{\lfloor sT \rfloor} \;=\;
    \sqrt{\alpha_{\lfloor sT \rfloor}} \,\mathbf{x}_0^{\mathrm{ref}}
    \;+\;
    \sqrt{1 - \alpha_{\lfloor sT \rfloor}}\;\boldsymbol{\epsilon},
    \label{eq:forward_i2i}
\end{equation}
with $\boldsymbol{\epsilon}$ drawn from a Gaussian distribution, and $\alpha_{\lfloor sT \rfloor}$ following a noise schedule. The diffusion model then performs the backward (denoising) steps from $t = \lfloor sT \rfloor$ down to $t=0$:
\begin{equation}
    \mathbf{x}_{t-1} \;=\;
    \mathbf{x}_{t}
    \;-\;
    \Delta\!\bigl(\mathbf{x}_{t},\,c,\,t;\,\theta\bigr),
    \quad t=\lfloor sT \rfloor,\dots,1,
    \label{eq:backward_i2i}
\end{equation}
where $\Delta(\cdot)$ represents the model-predicted noise removal and sampling correction term. 
The scalar $s$ is the translation strength: 
larger $s$ induces higher noise injection (more diverse outputs), 
while smaller $s$ preserves more details of $\mathbf{x}_0^{\mathrm{ref}}$~\cite{ho2020denoising,brooks2023instructpix2pix,mengsdedit,tumanyan2023plug,songdenoising}.

\subsection{Diffusion Model Fine-tuning}
Let $\epsilon_\theta(\mathbf{x}_t, c, t)$ be the noise predictor of a diffusion model, where $\mathbf{x}_t$ is a noisy sample at time step $t$, and $c$ is a text condition (e.g., a prompt). The simplified training objective commonly used is
\begin{equation}
    \mathcal{L}(\theta) \;=\; \mathbb{E}_{\mathbf{x}_0,\,t,\,\boldsymbol{\epsilon}}
    \Bigl\|\boldsymbol{\epsilon} \;-\; \epsilon_\theta(\mathbf{x}_t,\,c,\,t)\Bigr\|^2,
    \label{eq:diffusion_objective}
\end{equation}

\textbf{Textual Inversion.}~\cite{galimage}
Define a new learnable token embedding $\mathbf{v}$, appended to the text encoder vocabulary. 
During fine-tuning, we replace $c$ with $[\dots,\mathbf{v},\dots]$ in Eq.~\eqref{eq:diffusion_objective}, 
and optimize $\mathbf{v}$ such that the model captures the desired concept:
\begin{equation}
    \min_{\mathbf{v}} \;\mathbb{E}_{\mathbf{x}_0,\,t,\,\boldsymbol{\epsilon}}
    \Bigl\|\boldsymbol{\epsilon} - \epsilon_\theta(\mathbf{x}_t,\,c(\mathbf{v}),\,t)\Bigr\|^2,
    \label{eq:textual_inversion}
\end{equation}

\textbf{DreamBooth fine-tuning with LoRA.}
Instead of learning a textual embedding $\mathbf{v}$ alone, DreamBooth~\cite{ruiz2023dreambooth} fine-tunes a subset of model parameters 
$\phi \subset \theta$, specifically focusing on the U-Net~\cite{ronneberger2015u} of the diffusion model. 
To reduce the amount of parameters to be updated, LoRA (Low-Rank Adaptation)~\cite{hu2022lora} is commonly employed, 
which injects low-rank learnable matrices into the weight tensors of the U-Net.
Formally, let $\mathbf{W} \in \mathbb{R}^{d_{\mathrm{out}} \times d_{\mathrm{in}}}$ be a model weight, we decompose its update as
\begin{equation}
    \Delta \mathbf{W} = \mathbf{A} \mathbf{B},
    \quad \text{where}
    \quad \mathbf{A} \in \mathbb{R}^{d_{\mathrm{out}} \times r}, \quad 
    \mathbf{B} \in \mathbb{R}^{r \times d_{\mathrm{in}}},
\end{equation}
and $r \ll \min(d_{\mathrm{in}}, d_{\mathrm{out}})$.
During fine-tuning, we optimize only $\mathbf{A}, \mathbf{B}$, leaving the original $\mathbf{W}$ frozen. 
Thus, the effective trainable parameters are greatly reduced.

We formulate the overall training objective as:
\begin{equation}
    \min_{\phi} \;
    \mathbb{E}_{\mathbf{x}_0,\,t,\,\boldsymbol{\epsilon}}
    \Bigl\|\boldsymbol{\epsilon} - \epsilon_{\theta\setminus\phi,\phi}
    (\mathbf{x}_t,\,c,\,t)\Bigr\|^2,
    \label{eq:dreambooth_lora}
\end{equation}
where $\theta \setminus \phi$ denotes the frozen parameters and $\phi$ the low-rank adapted subset. 
By restricting the model updates to the LoRA components within the U-Net, 
we retain much of the original diffusion model's capacity while guiding it to learn 
the specific concept or style from limited reference images~\cite{kumari2023multi,patel2024conceptbed,qiu2023controlling}.

\begin{figure}[!t]
    \centering
    \includegraphics[width=\columnwidth]{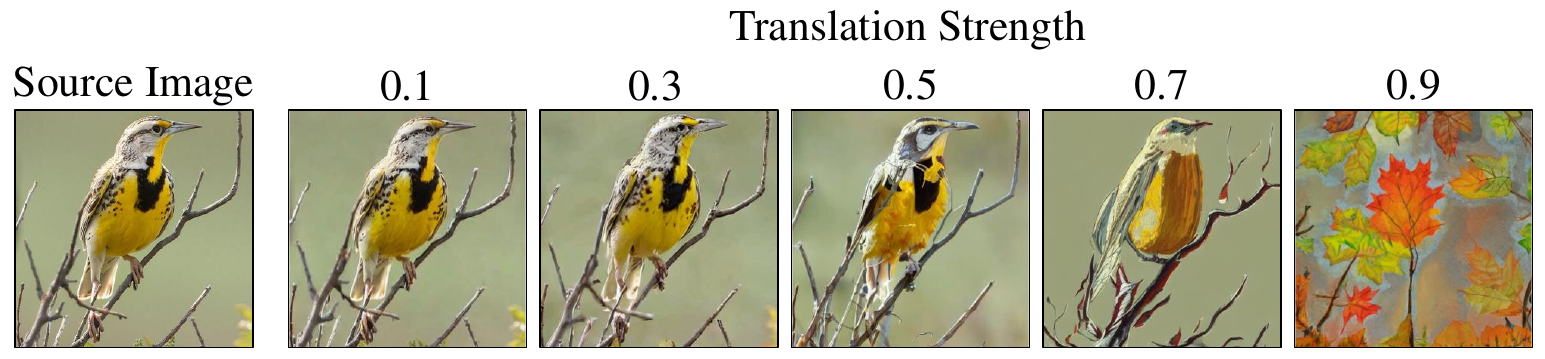} 

    \caption{Semantic drift in DiffuseMix~\cite{islam2024diffusemix} generated images using the prompt ``A transformed version of image into autumn''. Stronger transformations reduce semantic fidelity to the source image. More examples in \textcolor{magenta}{Supplementary Materials 2}.}
    \label{fig:diffusemix_semantic_drift}

\end{figure}

\section{Motivation}
\label{motivation}

Current SOTA methods, such as Diff-Mix~\cite{wang2024enhance} and DiffuseMix~\cite{islam2024diffusemix}, use diffusion models for data augmentation in domain-specific image classification. Yet, they struggle to balance diversity, faithfulness, and label clarity due to their designs not explicitly targeting these aspects and diffusion model limitations\footnote{See \textcolor{magenta}{Supplementary Materials 1 \& 2} for a detailed breakdown of these dependencies.}. Below, we briefly analyze these shortcomings to motivate our proposed SGD-Mix.

\begin{figure*}[!t]
    \centering
    \includegraphics[width=\textwidth]{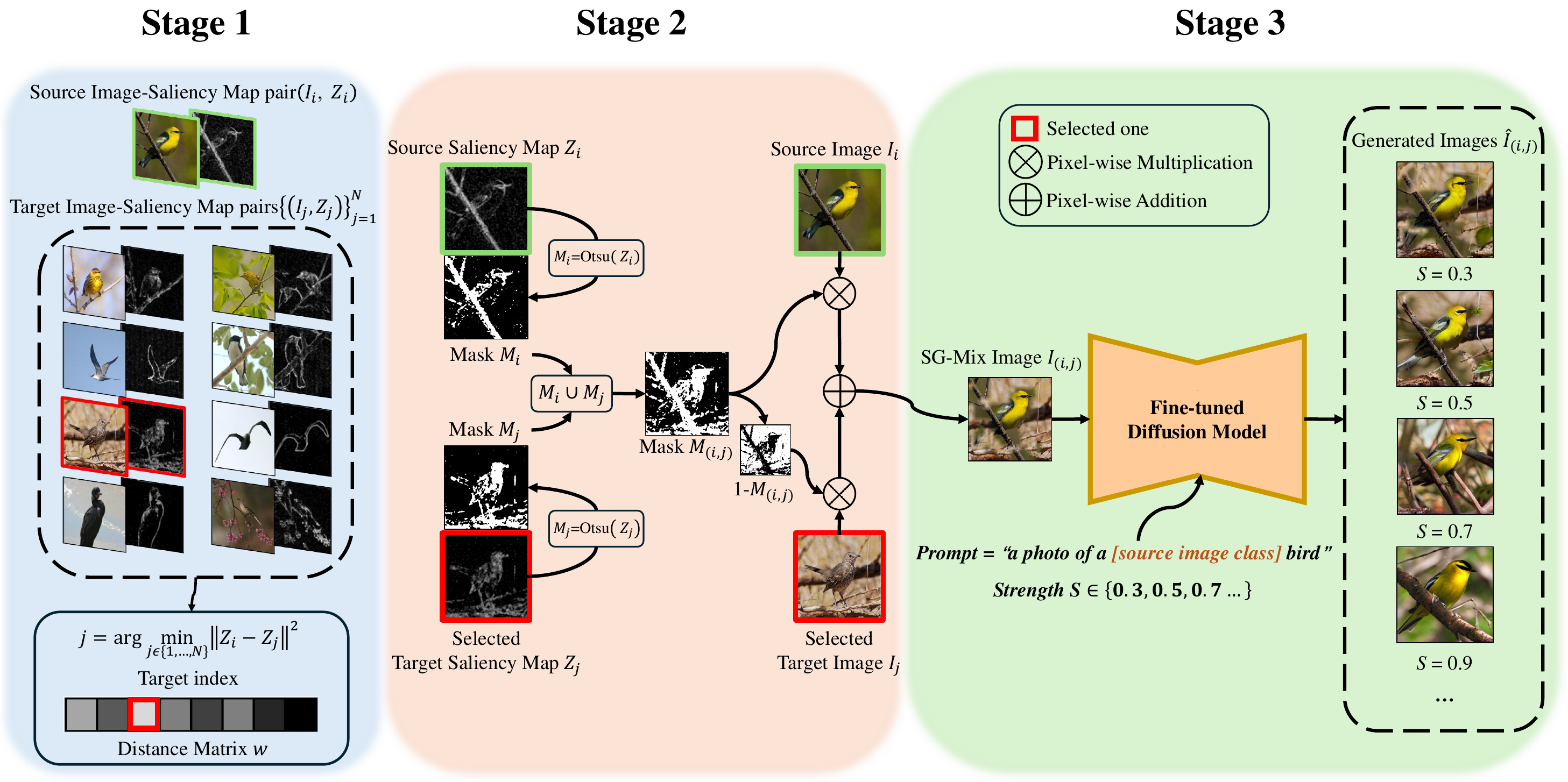}

    \caption{
    The pipeline of the proposed SGD-Mix method. 
    The process involves three major stages: (1) saliency-based target selection, (2) saliency-guided mixing, and (3) refining mixed images using our domain-specific fine-tuned diffusion model.}

    \label{fig:sgd-mix-pipeline}
\end{figure*}

\paragraph{Limitations of Diff-Mix.}
Diff-Mix generates inter-class images using a fine-tuned diffusion model and assigns labels via a nonlinear mixing formula, $\tilde{y} = (1 - s^\gamma) y^i + s^\gamma y^j$, where $s \in [0, 1]$ controls translation strength, $\gamma$ (often $\approx 0.5$) introduces nonlinearity, and $y^i$ and $y^j$ represent the labels of the reference (source) and target image classes, respectively. While effective in some cases, this approach has two key limitations. First, its label mixing lacks generalizability, as for a given $s$, the amount of information retained from the source image varies due to different characteristics of the diffusion model (e.g., noise scheduler~\cite{nichol2021improved}) and dataset-specific factors (e.g., image distribution), yet these images are assigned the same label regardless of such variations, affecting the consistency and reliability of the labeling process. Second, higher $s$ values increase stochasticity~\cite{mengsdedit}, yielding images with inconsistent semantics for the same $s$ that may mismatch their assigned labels. These issues stem from relying solely on $s$ for label determination, limiting robustness across different diffusion backbones and datasets. See Figure~\ref{fig:mixup_comparison} for an example where a Diff-Mix-generated image has a label composition over 0.8 for the source class and under 0.2 for the target class, which is clearly incorrect.

\paragraph{Limitations of DiffuseMix.}
In contrast to Diff-Mix’s focus on inter-class mixing, DiffuseMix transforms a source image with a conditional prompt, stylizing it via a pre-trained diffusion model without disentangling the foreground and background, then merging it with the original using a binary mask before blending with a fractal image for extra variation. Our experiments and findings from DEADiff~\cite{qi2024deadiff} indicate that while this approach is creative, it may suffer from semantic drift, where the stylized image deviates significantly from the source image's semantic content, particularly under strong transformations (see Figure~\ref{fig:diffusemix_semantic_drift}). Similar to CutMix~\cite{yun2019cutmix}, DiffuseMix also introduces unnatural boundaries due to the binary mask, which can further degrade the visual coherence of the generated image. Assigning the source image's label to the resulting image without label adjustment increases the risk of mislabeling. Moreover, retaining half of the source image in the mix limits diversity, thereby reducing the effectiveness of data augmentation.

Our analysis reveals that Diff-Mix and DiffuseMix are hindered by design and diffusion model limitations. Diff-Mix’s inter-class mixing leads to label ambiguity, while DiffuseMix’s thin transformations risk semantic drift and limited diversity. To address these issues, we propose SGD-Mix, a saliency-guided framework that, with a fine-tuned diffusion model, avoids inter-class mixing and mitigates semantic drift to ensure clear labels across the entire design, while also providing sufficient diversity and faithfulness for robust data augmentation.

\section{SGD-Mix}

The pipeline of SGD-Mix consists of three major stages: (1) selecting a target image and its saliency map based on the overlap of foreground regions, measured by the L2 distance between saliency maps, (2) generating a mixed image using saliency-based binary masks, and (3) refining the mixed image using our domain-specific fine-tuned diffusion model. The overall process is illustrated in Figure~\ref{fig:sgd-mix-pipeline}. The details of each stage are described below.

\subsection{Saliency-Based Target Selection}  
The purpose of this step is to find a target image whose foreground region overlaps the most with that of the source image, ensuring that the target image's background can be maximally retained in the subsequent steps. Given a source image $I_i$ and its normalized saliency map $Z_i$, we aim to select a target image $I_j$ and its corresponding normalized saliency map $Z_j$ from a target batch $\{(I_j, Z_j)\}_{j=1}^N$. The target batch is randomly sampled from the training dataset, where $N$ is a hyperparameter specifying the batch size. The selection criterion is based on the overlap of foreground regions, quantified using the L2 distance between saliency maps:
\begin{equation}
    j = \arg\min_{j \in \{1, \dots, N\}} \|Z_i - Z_j\|^2,
\end{equation}
where $j$ is the index of the selected target image. Saliency maps $Z$ can be generated using various existing methods, including rule-based~\cite{hou2007saliency} or gradient-based methods~\cite{simonyan2014deep,zhao2015saliency,selvaraju2017grad,zhou2016learning,huang2021snapmix}, and are all normalized to the range $[0, 1]$ using MinMax scaling.

\begin{figure}[!t]
    \centering
    \includegraphics[width=\columnwidth]{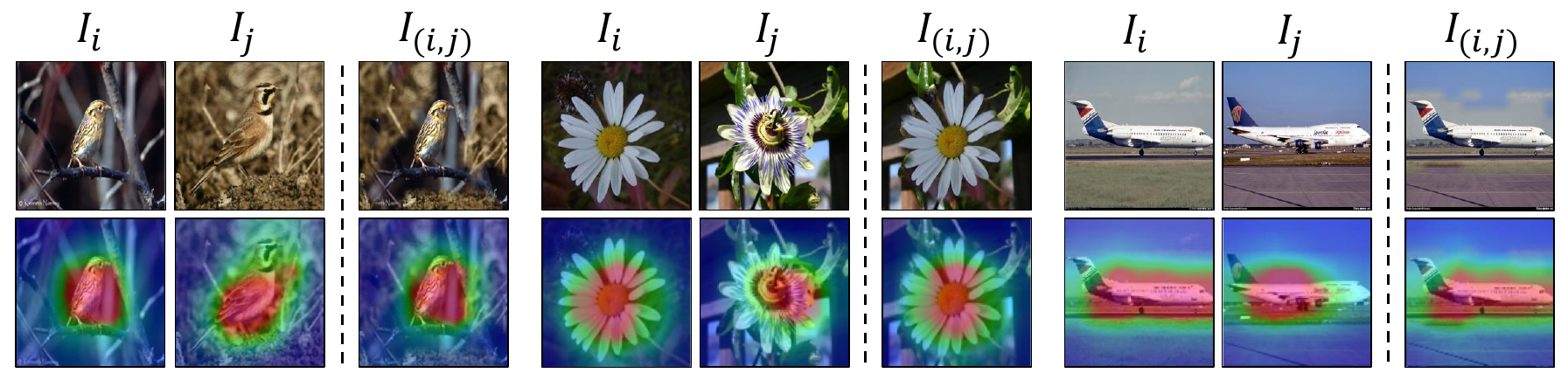}

    \caption{
    Visualization of attention maps before and after saliency-guided mixing in SGD-Mix. For a source image $I_i$ and target image $I_j$, the attention maps (bottom row) consistently focus on $I_i$'s foreground region in the mixed image $I_{(i,j)}$, preserving semantic consistency. More examples in \textcolor{magenta}{Supplementary Materials 4.1}.}

    \label{fig:sgdmix-attention-example}
\end{figure}

\subsection{Saliency-Guided Mixing}
The purpose of this step is to generate a mixed image where the foreground semantics are preserved from the source image while fully replacing the background with that of the target image. After identifying the target image $I_j$ and saliency map $Z_j$, binary masks are created for both the source and target saliency maps using Otsu's~\cite{otsu1975threshold} thresholding method:
\begin{equation}
    M_i = \text{Otsu}(Z_i), \quad M_j = \text{Otsu}(Z_j),
\end{equation}
This thresholding operation filters out the foreground parts of the images by segmenting the salient foreground from the less significant background, resulting in binary masks $M_i$ and $M_j$ that highlight the foreground regions. The masks are then combined using a union operation to form $M_{(i,j)}$:
\begin{equation}
    M_{(i,j)} = M_i \cup M_j,
\end{equation}
where $M_{(i,j)}$ prioritizes the retention of the source foreground while suppressing the target foreground interference. Using this mask, a saliency-guided mixed image $I_{(i,j)}$ is generated via pixel-wise composition:
\begin{equation}
    I_{(i,j)} = M_{(i,j)} \odot I_i + (1 - M_{(i,j)}) \odot I_j,
\end{equation}
where $\odot$ denotes pixel-wise multiplication. The mixed image retains the target image's background while ensuring that the foreground semantics exclusively originate from the source image.

The combined objective of the two steps is to completely replace the background of the source image with that of the target image, avoiding the introduction of foreign foreground semantics from other images. At the same time, the original foreground semantics of the source image are preserved, ensuring that the label of the mixed image remains consistent with that of the source image. This process is visualized in Figure~\ref{fig:sgdmix-attention-example}, which highlights how attention remains focused on the source image's foreground before and after mixing.

\subsection{Refining Mixed Images Using Our Domain-Specific Fine-Tuned Diffusion Model}  
To generate high-quality, class-specific images, it is feasible to directly fine-tune a diffusion model using DreamBooth~\cite{ruiz2023dreambooth} fine-tuning with LoRA~\cite{hu2022lora} on a domain-specific dataset. By providing class names as prompts (e.g., \texttt{``Prothonotary Warbler''}), the model can generate images corresponding to these classes. However, this approach faces significant challenges in datasets containing fine-grained categories with high inter-class similarity. For instance, species such as \texttt{``Prothonotary Warbler''} and \texttt{``Mourning Warbler''} exhibit substantial visual and semantic overlap. This similarity can hinder the model's ability to distinguish between closely related classes, leading to convergence issues and reduced specificity in generated outputs.

\begin{figure}[t]
    \centering
    \includegraphics[width=\columnwidth]{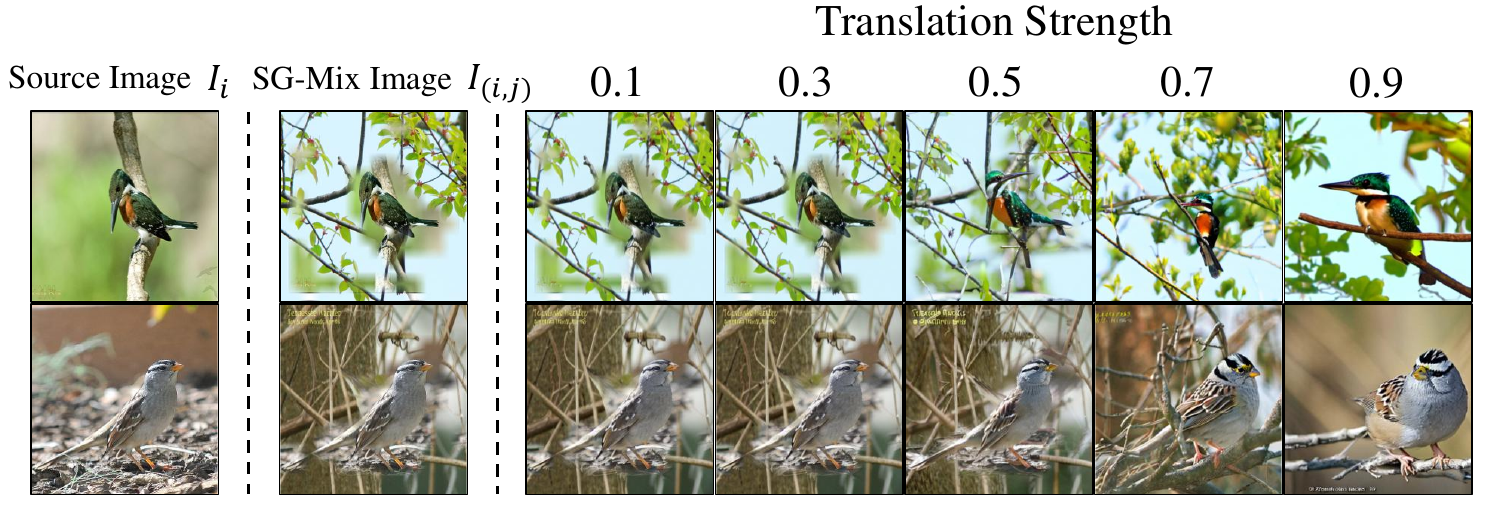}

    \caption{
    Examples of SGD-Mix generated images under varying translation strengths \( S \in \{0.1, 0.3, 0.5, 0.7, 0.9\} \). The generated image retains the source image's foreground semantics while the background evolves with increasing \( S \), balancing diversity and faithfulness. More examples in \textcolor{magenta}{Supplementary Materials 4.2}.}

    \label{fig:sgdmix-generation-example}
\end{figure}

To address these challenges, we draw inspiration from Diff-Mix~\cite{wang2024enhance} and adopt a structured approach that combines DreamBooth fine-tuning with LoRA and Textual Inversion~\cite{galimage}. First, we replace the direct usage of class names with structured identifiers to disentangle fine-grained semantics. Specifically, the class names are reformulated as: \text{``[$v^i$] [$metaclass$]''}, where \( v^i \) is a learnable embedding representing the \( i \)-th class (\( i \in \{1, 2, \dots, n\} \)), and \( [metaclass] \) describes the overarching category (e.g., ``bird''). DreamBooth fine-tuning with LoRA is then utilized to efficiently adapt the diffusion model's U-Net~\cite{ronneberger2015u} to the domain-specific distributions, enabling the model to generate images that align with the target dataset's visual characteristics.

This joint fine-tuning process optimizes:
\begin{itemize}
    \item \textbf{Learnable identifiers ($v^i$)}: These embeddings provide compact and distinct representations for each fine-grained category, ensuring effective semantic disentanglement.
    \item \textbf{Low-Rank Adaptations (LoRA)}: Efficiently adjust the model parameters to capture domain-specific nuances.
\end{itemize}

During inference, the fine-tuned diffusion model utilizes the mixed image \( I_{(i,j)} \) as a conditioning input and generates images guided by a structured prompt:
\[
\text{``a photo of a [$v^s$] [$metaclass$]''},
\]
where \( v^s \) is the learnable embedding corresponding to the source image class. This structured prompt ensures that the generated image aligns with the semantic content of the source image while preserving the saliency of the source foreground and incorporating the background diversity introduced in previous steps.

Finally, the label of the generated image is determined based on our label-preserving process. Since the mixed image \( I_{(i,j)} \) inherits its label from the source image \( I_i \), and the diffusion model generates outputs conditioned solely on \( v^s \) without introducing semantics from other classes, the label of the generated image remains consistent with that of the source image which ensures that our method achieves label clarity:
\begin{equation}
\text{Label}(\hat{I}_{(i,j)}) = \text{Label}(I_{(i,j)}) = \text{Label}(I_i).
\end{equation}

To further enhance flexibility, we introduce a Strength parameter \( S \), which controls the noise injection level during image generation. By adjusting \( S \), users can balance faithfulness to the source image with diversity in the generated outputs. This mechanism ensures that the refined images achieve the desired levels of faithfulness, diversity, and label clarity, making them suitable for downstream tasks. The effect of varying \( S \) on the generated images is demonstrated in Figure~\ref{fig:sgdmix-generation-example}.

\begin{algorithm}[t]
\caption{SGD-Mix}
\label{alg:sgdmix}
\textbf{Input:} A single training dataset of $m$ image-saliency pairs $\{(I_k, Z_k)\}_{k=1}^m$, Batch size $N$, Domain-specific fine-tuned diffusion model $\mathcal{D}$

\textbf{Output:} Generated image set $\hat{\mathcal{I}}$

\begin{algorithmic}[1]
\State Initialize $\hat{\mathcal{I}} \gets \emptyset$
\For {$i = 1$ to $m$}
    \State \textbf{Sample Target Batch (from the same dataset):}
    \Statex \hspace{2em} \parbox[t]{0.85\linewidth}{Randomly sample $\{(I_j, Z_j)\}_{j=1}^N$ from $\{(I_k, Z_k)\}_{k=1}^m \setminus \{(I_i, Z_i)\}$}
    
    \State \textbf{Target Selection:} 
    \Statex \hspace{2em} $j \gets \arg\min\limits_{j \in \{1,\dots,N\}} \|Z_i - Z_j\|^2$
    
    \State \textbf{Mask Generation:} 
    \Statex \hspace{2em} \parbox[t]{0.85\linewidth}{Compute $M_i \gets \mathrm{Otsu}(Z_i)$ and $M_j \gets \mathrm{Otsu}(Z_j)$}
    \Statex \hspace{2em} Merge masks: $M_{(i,j)} \gets M_i \cup M_j$
    
    \State \textbf{Saliency-Guided Mixing:} 
    \Statex \hspace{2em} \parbox[t]{0.85\linewidth}{Compute mixed image: $I_{(i,j)} \gets M_{(i,j)} \odot I_i + (1 - M_{(i,j)}) \odot I_j$}
    
    \State \textbf{Diffusion-Based Refinement:} 
    \Statex \hspace{2em} \parbox[t]{0.85\linewidth}{Generate final image: $\hat{I}_{(i,j)} \gets \mathcal{D}(I_{(i,j)})$}
    \Statex \hspace{2em} Update generated set: $\hat{\mathcal{I}} \gets \hat{\mathcal{I}} \cup \{\hat{I}_{(i,j)}\}$
\EndFor
\State \textbf{return} $\hat{\mathcal{I}}$
\end{algorithmic}
\end{algorithm}

\section{Experiments}
\label{experiments}

\begin{table*}[t]
\centering
\footnotesize
\resizebox{\textwidth}{!}{
\begin{tabular}{l|cccccc|cccccc}
\toprule
\multirow{2}{*}{Method} & \multicolumn{6}{c|}{\textbf{ResNet50@448}} & \multicolumn{6}{c}{\textbf{ViT-B/16@384}} \\
\cmidrule(lr){2-7} \cmidrule(lr){8-13}
& CUB & Aircraft & Flower & Car & Dog & Avg & CUB & Aircraft & Flower & Car & Dog & Avg \\
\midrule
- & 86.64 & 89.09 & 99.27 & 94.54 & 87.48 & 91.40 & 89.37 & 83.50 & 99.56 & 94.21 & 92.06 & 91.74 \\
CutMix~\cite{yun2019cutmix} & \uline{87.23} & 89.44 & 99.25 & 94.73 & 87.59 & 91.65 & \textbf{90.52} & 83.50 & 99.64 & 94.83 & \uline{92.13} & 92.12 \\
Mixup~\cite{zhang2018mixup} & 86.68 & 89.41 & 99.40 & 94.49 & 87.42 & 91.48 & 90.32 & 84.31 & \textbf{99.73} & 94.98 & 92.02 & \uline{92.27} \\
GuidedMixup~\cite{kang2023guidedmixup}$\ddagger$ & 86.50 & 88.60 & 99.49 & 94.99 & 87.52 & 91.42 & 90.00 & 82.99 & 98.98 & 94.20 & \textbf{92.51} & 91.70 \\
Real-Filtering~\cite{hesynthetic} & 85.60 & 88.54 & 99.09 & 94.59 & 87.30 & 91.22 & 89.49 & 83.07 & 99.36 & 94.66 & 91.91 & 91.69 \\
Real-Guidance~\cite{hesynthetic} & 86.71 & 89.07 & 99.25 & 94.55 & 87.40 & 91.59 & 89.54 & 83.17 & 99.59 & 94.65 & 92.05 & 91.80 \\
Da-Fusion~\cite{trabuccoeffective} & 86.30 & 87.64 & 99.37 & 94.69 & 87.33 & 91.07 & 89.40 & 81.88 & 99.61 & 94.53 & 92.07 & 91.50 \\
Diff-Mix~\cite{wang2024enhance} & 87.16 & \textbf{90.25} & \uline{99.54} & \uline{95.12} & \uline{87.74} & \uline{91.96} & 90.05 & \uline{84.33} & \uline{99.64} & \uline{95.09} & 91.99 & 92.22 \\

DiffuseMix~\cite{islam2024diffusemix}$\ddagger$ & 86.19 & 88.81 & 99.08 & 94.59 & 87.39 & 91.21 & 88.99 & 83.02 & 99.20 & 94.17 & 91.63 & 91.40 \\
\rowcolor[gray]{0.9} \textbf{SGD-Mix (Ours)} & \textbf{87.66} & \uline{90.16} & \textbf{99.59} & \textbf{95.27} & \textbf{88.01} & \textbf{92.14} & \uline{90.44} & \textbf{85.21} & 99.61 & \textbf{95.32} & 91.99 & \textbf{92.51} \\
\bottomrule
\end{tabular}}
\caption{Fine-grained classification accuracy (\%).}
\label{fgvc-results}
\end{table*}

\begin{table*}[t]
\centering
\footnotesize
\resizebox{\textwidth}{!}{
\begin{tabular}{l|cccc|cc|cccc|cc}
\toprule
\multirow{2}{*}{Method} & \multicolumn{6}{c|}{\textbf{CUB-LT}} & \multicolumn{6}{c}{\textbf{Flower-LT}} \\
\cmidrule(lr){2-7} \cmidrule(lr){8-13}
& \multicolumn{4}{c|}{\textbf{IF=100}} & \textbf{50} & \textbf{10} & \multicolumn{4}{c|}{\textbf{IF=100}} & \textbf{50} & \textbf{10} \\
& Many & Medium & Few & All & All & All & Many & Medium & Few & All & All & All \\
\midrule
- & 79.11 & 64.28 & 13.48 & 33.65 & 44.82 & 58.13 & 99.19 & 94.95 & 58.18 & 80.43 & 90.87 & 95.07 \\
CMO~\cite{park2022majority} & 78.32 & 58.57 & 14.78 & 32.94 & 44.08 & 57.62 & 99.25 & 95.19 & 67.45 & 83.95 & 91.43 & 95.19 \\
CMO+DRW~\cite{cao2019learning} & 78.97 & 56.36 & 14.66 & 32.57 & 46.43 & 59.25 & \textbf{99.97} & 95.06 & 67.31 & 84.07 & 92.06 & 95.92 \\
Copy-Paste (LSJ)~\cite{ghiasi2021simple}$\ddagger$ & 78.55 & 58.48 & 11.99 & 30.95 & 43.80 & 57.92 & 98.73 & 93.02 & 57.00 & 75.89 & 89.02 & 94.97 \\
SaliencyMix~\cite{uddinsaliencymix}$\ddagger$ & 81.93 & 62.05 & 14.29 & 33.71 & 46.03 & 61.01 & 99.04 & 93.30 & 58.36 & 76.71 & 89.90 & 95.02 \\
DA-fusion~\cite{trabuccoeffective}$\ddagger$ & 81.20 & 61.99 & 27.50 & 41.87 & 49.55 & 59.98 & 99.20 & 94.97 & 64.04 & 80.16 & 91.02 & 94.97 \\
Real-Gen~\cite{wang2024enhance} & \textbf{84.88} & 65.23 & 30.68 & 45.86 & 53.43 & 61.42 & 98.64 & 95.55 & 66.10 & 83.56 & 91.84 & 95.22 \\
Real-Mix~\cite{wang2024enhance} & \uline{84.63} & 66.34 & 34.44 & 47.75 & 55.67 & 62.27 & \uline{99.87} & 96.26 & 68.53 & 85.19 & 92.96 & 96.04 \\
Diff-Mix~\cite{wang2024enhance} & 84.07 & \textbf{67.79} & \uline{36.55} & \textbf{50.35} & \uline{58.19} & \uline{64.48} & 99.25 & \textbf{96.98} & \uline{78.41} & \textbf{89.46} & \uline{93.86} & \uline{96.63} \\
DiffuseMix~\cite{islam2024diffusemix}$\ddagger$ & 80.96 & 63.97 & 31.02 & 44.65 & 51.98 & 59.87 & 99.04 & 95.32 & 65.94 & 81.20 & 90.94 & 95.99 \\
\rowcolor[gray]{0.9} \textbf{SGD-Mix (Ours)} & 84.34 & \uline{66.80} & \textbf{37.00} & \uline{49.48} & \textbf{58.68} & \textbf{64.98} & 99.68 & \uline{96.30} & \textbf{79.98} & \uline{88.66} & \textbf{94.03} & \textbf{96.89} \\
\bottomrule
\end{tabular}}
\caption{Long-tail classification accuracy (\%).}
\label{tab:cub_flower_lt}
\end{table*}

We evaluate our proposed SGD-Mix method through four key studies: (1) fine-grained vision classification, (2) long-tail classification, (3) few-shot classification, and (4) background robustness. These assess SGD-Mix under diverse challenges—high-resolution inputs, imbalanced distributions, data scarcity, and background shifts—comparing it against SOTA data augmentation methods to highlight its effectiveness in domain-specific image classification. Unless specified, we set batch size $N=50$, compute saliency maps via gradient-based method (Grad-CAM)~\cite{selvaraju2017grad}, and follow standard protocols for fair comparisons\footnote{Results marked with $\ddagger$ in the tables are our reproductions; others are reported by Diff-Mix~\cite{wang2024enhance}, slightly outperforming ours under consistent settings. We adopt their values for fairness and rigor.}. More details of the experimental setup are in the \textcolor{magenta}{Supplementary Materials 3}.

\subsection{Fine-Grained Vision Classification}
\label{fgvc-experiments}

Fine-grained visual classification (FGVC) involves distinguishing subtle intra-category differences. We evaluate SGD-Mix on five datasets—CUB~\cite{wah2011caltech}, Stanford Cars~\cite{krause20133d}, Oxford Flowers~\cite{nilsback2008automated}, Stanford Dogs~\cite{khosla2011novel}, and FGVC Aircraft~\cite{maji2013fine}—using ResNet50~\cite{he2016deep} (ImageNet1K-pretrained~\cite{russakovsky2015imagenet}, $448\times448$) and ViT-B/16~\cite{dosovitskiy2020image} (ImageNet21K-pretrained~\cite{ridnik1imagenet}, $384\times384$). SGD-Mix employs a translation strength of $S \in \{0.5, 0.7, 0.9\}$, label smoothing~\cite{muller2019does} with confidence 0.9, an expansion multiplier of 5, and a replacement probability of $p=0.1$. Competitors include generative diffusion-based methods (Diff-Mix~\cite{wang2024enhance}, DiffuseMix~\cite{islam2024diffusemix}, Real-Filtering~\cite{hesynthetic}, Real-Guidance~\cite{hesynthetic}, Da-Fusion~\cite{trabuccoeffective}) and non-generative methods (Mixup~\cite{zhang2018mixup}, CutMix~\cite{yun2019cutmix}, GuidedMixup~\cite{kang2023guidedmixup}). For fair comparison, Diff-Mix and DiffuseMix adopt the same $S$ as SGD-Mix, with Diff-Mix using $\gamma=0.5$, while Da-Fusion uses a randomly sampled $S \in \{0.25, 0.5, 0.75, 1.0\}$.

Table~\ref{fgvc-results} shows SGD-Mix outperforming SOTA, achieving the highest average accuracy (92.14\% ResNet50, 92.51\% ViT) across datasets, surpassing Diff-Mix (91.96\% ResNet50, 92.22\% ViT) by +0.18\% and +0.29\%, with notable gains on Dog (+0.27\% ResNet50) and Aircraft (+0.88\% ViT). \textit{SGD-Mix succeeds by generating data with reliable foregrounds and diverse backgrounds, capturing subtle class distinctions. See Q1 in Section~\ref{discussion} for details.}

\subsection{Long-Tail Classification}

Long-tail classification addresses class imbalance, a common real-world challenge. We test SGD-Mix on CUB-LT and Flower-LT datasets~\cite{samuel2021generalized,cao2019learning,liu2019large,park2022majority}, varying imbalance factors (IF=100, 50, 10), using SYNAuG~\cite{ye2023exploiting} to balance distributions with synthetic data. Competitors include generative diffusion-based methods (DiffuseMix~\cite{islam2024diffusemix}, Diff-Mix~\cite{wang2024enhance}, Real-Mix~\cite{wang2024enhance}, Real-Gen~\cite{wang2024enhance}, DA-fusion~\cite{trabuccoeffective}) and non-generative methods (SaliencyMix~\cite{uddinsaliencymix}, Copy-Paste~\cite{ghiasi2021simple}, CMO~\cite{park2022majority}, CMO+DRW~\cite{cao2019learning}), with $S=0.7$ for SGD-Mix/Diff-Mix/Real-Mix and $\gamma=0.5$ for Diff-Mix/Real-Mix.

Table~\ref{tab:cub_flower_lt} reveals SGD-Mix surpassing SOTA Diff-Mix on CUB-LT (64.98\% vs. 64.48\% at IF=10) and Flower-LT (96.89\% vs. 96.63\%), excelling in Few classes (+0.45\%/+1.57\% on CUB-LT/Flower-LT IF=100). \textit{Reliable intra-class foregrounds in reference images ensure accurate augmentation for rare classes. See Q1 in Section~\ref{discussion} for details.}

\begin{table}[t]
\centering
\footnotesize
\resizebox{\columnwidth}{!}{
\begin{tabular}{l|c|c|c|c}
\toprule
Method & 1-shot ($p=0.5$) & 5-shot ($p=0.3$) & 10-shot ($p=0.2$) & All-shot ($p=0.1$) \\
\midrule
-$\ddagger$ & 16.10 & 50.98 & 68.69 & 81.74 \\
Copy-Paste (LSJ)~\cite{ghiasi2021simple}$\ddagger$ & 16.81 & 52.97 & 69.02 & 81.77 \\
SaliencyMix~\cite{uddinsaliencymix}$\ddagger$ & 17.47 & 53.02 & 69.50 & 81.90 \\
DA-fusion~\cite{trabuccoeffective}$\ddagger$ & 20.02 & 53.04 & 69.97 & 81.79 \\
Diff-Aug~\cite{wang2024enhance}$\ddagger$ & 20.50 & 57.49 & 71.00 & 82.02 \\
Diff-Mix~\cite{wang2024enhance}$\ddagger$ & 25.49 & \uline{59.30} & \uline{71.51} & \uline{82.34} \\
Diff-Gen~\cite{wang2024enhance}$\ddagger$ & \textbf{26.10} & 58.35 & 71.14 & 82.26 \\
\rowcolor[gray]{0.9} \textbf{SGD-Mix (Ours)} & \uline{25.68} & \textbf{59.61} & \textbf{71.90} & \textbf{82.88} \\
\bottomrule
\end{tabular}}
\caption{Few-shot classification accuracy (\%).}
\label{tab:fewshot_results}
\end{table}

\subsection{Few-Shot Classification}

Few-shot classification tests generalization from limited data. On CUB~\cite{wah2011caltech} with ResNet50~\cite{he2016deep} ($224 \times 224$), we evaluate 1-shot, 5-shot, 10-shot, and all-shot settings, setting $S = 0.9$ for SGD-Mix, Diff-Mix~\cite{wang2024enhance}, and Diff-Aug~\cite{wang2024enhance}, with an expansion multiplier of 5 and replacement probability $p = \{0.5, 0.3, 0.2, 0.1\}$ respectively.

Table~\ref{tab:fewshot_results} shows SGD-Mix leading in 5-shot (59.61\% vs. 59.30\% Diff-Mix), 10-shot (71.90\% vs. 71.51\%), and all-shot (82.88\% vs. 82.34\%), with competitive 1-shot performance (25.68\% vs. 26.10\% Diff-Gen~\cite{wang2024enhance}). \textit{Reliable intra-class foregrounds in reference images ensure correct foreground generation despite limited samples. See Q1 in Section~\ref{discussion} for details.}

\subsection{Background Robustness}

To assess if diverse samples boost Diff-Mix’s resilience to background shifts, we test it on the out-of-distribution Waterbird dataset~\cite{sagawadistributionally} (CUB~\cite{wah2011caltech} foregrounds with Places~\cite{zhou2017places} backgrounds), split into (waterbird, water), (waterbird, land), (landbird, land), and (landbird, water). Models trained on CUB and synthetic data are evaluated for robustness.

Table~\ref{tab:waterbird_results} indicates SGD-Mix slightly edges out Diff-Mix (72.49\% vs. 72.47\% avg.), with a standout +6.5\% gain in (waterbird, land) over the baseline. \textit{Its foreground-background disentanglement mitigates background bias. See Q1 in Section~\ref{discussion} for details.}

\begin{table}[t]
\centering
\footnotesize
\resizebox{\columnwidth}{!}{
\begin{tabular}{l|c|c|c|c|c}
\toprule
Method & (waterbird, water) & (waterbird, land) & (landbird, land) & (landbird, water) & Avg \\
\midrule
- & 59.50 & 56.70 & 73.48 & 73.97 & 70.19 \\
CutMix~\cite{yun2019cutmix} & 62.46 & 60.12 & 73.39 & \textbf{74.72} & 71.23 \\
Copy-Paste (LSJ)~\cite{ghiasi2021simple}$\ddagger$ & 59.97 & 56.70 & 73.17 & 72.95 & 69.80 \\
SaliencyMix~\cite{uddinsaliencymix}$\ddagger$ & 62.15 & 60.28 & 73.48 & 73.53 & 70.78 \\
DA-Fusion~\cite{trabuccoeffective} & 60.90 & 58.10 & 72.94 & 72.77 & 69.90 \\
Diff-Aug~\cite{wang2024enhance} & 61.83 & 60.12 & 73.04 & 73.52 & 70.28 \\
Diff-Mix~\cite{wang2024enhance} & \uline{63.83} & \uline{63.24} & \textbf{75.64} & 74.36 & \uline{72.47} \\
\rowcolor[gray]{0.9} \textbf{SGD-Mix (Ours)} & \textbf{63.86} & \textbf{63.24} & \uline{75.57} & \uline{74.50} & \textbf{72.49} \\
\bottomrule
\end{tabular}}
\caption{Classification accuracy (\%) on the Waterbird dataset.}
\label{tab:waterbird_results}
\end{table}

\begin{table}[t]
\centering
\footnotesize
\begin{tabular}{l|c|c|c}
\toprule
Dataset & Baseline & SGD-Mix (SR) & SGD-Mix \\
\midrule
CUB & 86.64  & \uline{86.73} & \textbf{87.66} \\
Aircraft & 89.09 & \uline{89.11} & \textbf{90.16} \\
Flower & 99.27 & \uline{99.49} & \textbf{99.59} \\
Car & 94.54 & \uline{94.70} & \textbf{95.27} \\
Dog & \uline{87.48} & 86.96 & \textbf{88.01} \\
\bottomrule
\end{tabular}
\caption{Comparison of Baseline, SGD-Mix, and SGD-Mix (SR) using ResNet50 at 448$\times$448, per Section~\ref{fgvc-experiments}. SGD-Mix (SR) uses spectral residual~\cite{hou2007saliency} saliency maps.}
\label{tab:sgd_mix_sr}
\end{table}

\section{Discussion}
\label{discussion}

\paragraph{Q1: Why does SGD-Mix consistently outperform across diverse tasks?}
\label{discussion:q1}

\textbf{A:} SGD-Mix excels by jointly optimizing diversity, faithfulness, and label clarity—key factors for effective data augmentation in domain-specific image classification. It leverages saliency maps to separate foreground and background, preserving the source image’s foreground semantics while diversifying backgrounds with target images. A fine-tuned diffusion model further refines these mixes, generating high-quality outputs through non-linear synthesis. Unlike Diff-Mix~\cite{wang2024enhance}, which risks label ambiguity due to inter-class mixing and requires filters~\cite{radford2021learning} for unnatural images, SGD-Mix’s label-preserving method ensures semantic alignment, eliminating mismatches without the need for additional filtering. Similarly, unlike DiffuseMix~\cite{islam2024diffusemix}, our approach does not suffer from semantic drift. It follows the foreground-background disentanglement principle~\cite{wang2024enhance}, producing reliable foregrounds with diverse backgrounds to enhance robust feature learning. Moreover, its saliency-guided mixing inherently aligns with the VRM~\cite{chapelle2000vicinal}. These combined advantages strengthen robustness in fine-grained classification (capturing subtle details), long-tail settings (handling rare classes), few-shot learning (adapting to scarce data), and background robustness (mitigating bias).

\begin{figure}[t]
    \centering
    \begin{tikzpicture}
        \begin{axis}[
            width=\linewidth,
            height=7.0cm,
            xlabel={Batch Size $N$},
            xlabel style={font=\footnotesize},
            ylabel={Accuracy on CUB (\%)},
            ylabel style={font=\footnotesize},
            xmin=0, xmax=110,
            ymin=86, ymax=88.5,
            xtick={10, 20, 30, 50, 100},
            ytick={86, 87, 88},
            grid=major,
            legend pos=north west
        ]
        \addplot[color=red,mark=square*] coordinates {
            (10, 86.80)
            (20, 87.19)
            (30, 87.45)
            (50, 87.66)
            (100, 87.81)
        };
        \addlegendentry{SGD-Mix}
        \addplot[dashed,color=black,mark=*] coordinates {
            (10, 86.64)
            (20, 86.64)
            (30, 86.64)
            (50, 86.64)
            (100, 86.64)
        };
        \addlegendentry{Baseline}
        \end{axis}
    \end{tikzpicture}
    \caption{Accuracy on CUB with ResNet50, varying $N$ (Section~\ref{fgvc-experiments}). SGD-Mix improves with $N$, saturating near 50.}
    \label{fig:batch_size_effect}

\end{figure}
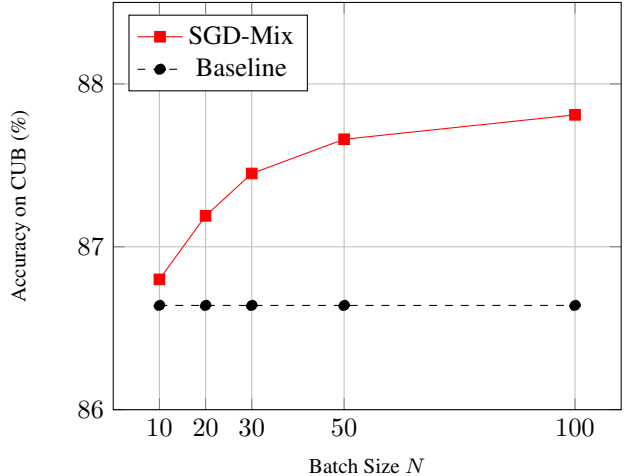

\paragraph{Q2: Why avoid additional background datasets or segmentation models, and how does SGD-Mix differ from simple background swapping?}
\label{discussion:q2}

\textbf{A:} We prioritize a method using only the input dataset, as external background datasets may cause semantic inconsistency due to uncertain contextual relevance. Segmentation models, though viable, are computationally costly and less adaptable, struggling to match foregrounds and backgrounds within a dataset without an optimal pairing mechanism. In contrast, SGD-Mix uses L2 distance between saliency maps to pair similar foregrounds with diverse backgrounds efficiently and precisely in a lightweight framework. Unlike simple background swapping~\cite{rother2004grabcut,yu2018generative,sengupta2020background}, which only swaps backgrounds without changing foregrounds and thus lacks diversity, SGD-Mix preserves foreground semantics, adds diverse backgrounds, and leverages a fine-tuned diffusion model for nonlinear and high-fidelity generation, overcoming these limitations.

\paragraph{Q3: Is SGD-Mix tied to gradient-based saliency maps?}
\label{discussion:q3}

\textbf{A:} No, SGD-Mix supports any saliency map method. While gradient-based (Grad-CAM)~\cite{selvaraju2017grad} saliency maps yield optimal accuracy, ablation studies (Table~\ref{tab:sgd_mix_sr}) with spectral residual methods~\cite{hou2007saliency} show minor performance drops yet sustained effectiveness, proving the framework’s flexibility.

\paragraph{Q4: How does batch size $N$ impact performance?}
\label{discussion:q4}

\textbf{A:} Ablation studies (Figure~\ref{fig:batch_size_effect}) testing $N = 10, 20, 30, 50, 100$ reveal: 1) small $N$ limits target selection, reducing diversity; 2) larger $N$ boosts augmentation quality, but increases computation (See \textcolor{magenta}{Supplementary Materials 3.3.2} for details.), plateauing beyond 50 with diminishing returns. We recommend \( N \in [30, 50] \) for efficiency and effectiveness.

\section{Conclusion}
\label{conclusion}

In this work, we introduce SGD-Mix, a data augmentation framework for domain-specific image classification that simultaneously enhances diversity, faithfulness, and label clarity in generated data. Using saliency-guided mixing and a fine-tuned diffusion model, SGD-Mix overcomes prior method limitations and diffusion model challenges. Experiments on fine-grained, long-tail, few-shot, and background robustness tasks show it outperforms state-of-the-art methods with balanced, robust augmentation.

\paragraph{Limitations}
SGD-Mix incurs computational overhead from saliency map processing, which may hinder scalability in resource-limited settings. We detail this in \textcolor{magenta}{Supplementary Materials 3.3.2} and plan to explore lightweight saliency processing techniques in future work.

{\small
\bibliographystyle{ieee}
\bibliography{main}

\begin{thebibliography}{10}\itemsep=-1pt

\bibitem{antoniou2018augmenting}
A.~Antoniou, A.~Storkey, and H.~Edwards.
\newblock Augmenting image classifiers using data augmentation generative
  adversarial networks.
\newblock In {\em International conference on artificial neural networks},
  pages 594--603. Springer, 2018.

\bibitem{antoniou2018data}
A.~Antoniou, A.~Storkey, and H.~Edwards.
\newblock Data augmentation generative adversarial networks.
\newblock {\em stat}, 1050:8, 2018.

\bibitem{araslanov2021self}
N.~Araslanov and S.~Roth.
\newblock Self-supervised augmentation consistency for adapting semantic
  segmentation.
\newblock In {\em Proceedings of the IEEE/CVF conference on computer vision and
  pattern recognition}, pages 15384--15394, 2021.

\bibitem{azizisynthetic}
S.~Azizi, S.~Kornblith, C.~Saharia, M.~Norouzi, and D.~J. Fleet.
\newblock Synthetic data from diffusion models improves imagenet
  classification.
\newblock {\em Transactions on Machine Learning Research}.

\bibitem{brooks2023instructpix2pix}
T.~Brooks, A.~Holynski, and A.~A. Efros.
\newblock Instructpix2pix: Learning to follow image editing instructions.
\newblock In {\em Proceedings of the IEEE/CVF conference on computer vision and
  pattern recognition}, pages 18392--18402, 2023.

\bibitem{cao2019learning}
K.~Cao, C.~Wei, A.~Gaidon, N.~Arechiga, and T.~Ma.
\newblock Learning imbalanced datasets with label-distribution-aware margin
  loss.
\newblock {\em Advances in neural information processing systems}, 32, 2019.

\bibitem{chapelle2000vicinal}
O.~Chapelle, J.~Weston, L.~Bottou, and V.~Vapnik.
\newblock Vicinal risk minimization.
\newblock {\em Advances in neural information processing systems}, 13, 2000.

\bibitem{chen2023importance}
T.~Chen.
\newblock On the importance of noise scheduling for diffusion models.
\newblock {\em arXiv preprint arXiv:2301.10972}, 2023.

\bibitem{dhariwal2021diffusion}
P.~Dhariwal and A.~Nichol.
\newblock Diffusion models beat gans on image synthesis.
\newblock {\em Advances in neural information processing systems},
  34:8780--8794, 2021.

\bibitem{dosovitskiy2020image}
A.~Dosovitskiy, L.~Beyer, A.~Kolesnikov, D.~Weissenborn, X.~Zhai,
  T.~Unterthiner, M.~Dehghani, M.~Minderer, G.~Heigold, S.~Gelly, et~al.
\newblock An image is worth 16x16 words: Transformers for image recognition at
  scale.
\newblock In {\em International Conference on Learning Representations}, 2020.

\bibitem{galimage}
R.~Gal, Y.~Alaluf, Y.~Atzmon, O.~Patashnik, A.~H. Bermano, G.~Chechik, and
  D.~Cohen-or.
\newblock An image is worth one word: Personalizing text-to-image generation
  using textual inversion.
\newblock In {\em The Eleventh International Conference on Learning
  Representations}.

\bibitem{ghiasi2021simple}
G.~Ghiasi, Y.~Cui, A.~Srinivas, R.~Qian, T.-Y. Lin, E.~D. Cubuk, Q.~V. Le, and
  B.~Zoph.
\newblock Simple copy-paste is a strong data augmentation method for instance
  segmentation.
\newblock In {\em Proceedings of the IEEE/CVF conference on computer vision and
  pattern recognition}, pages 2918--2928, 2021.

\bibitem{goodfellow2014generative}
I.~Goodfellow, J.~Pouget-Abadie, M.~Mirza, B.~Xu, D.~Warde-Farley, S.~Ozair,
  A.~Courville, and Y.~Bengio.
\newblock Generative adversarial nets.
\newblock {\em Advances in neural information processing systems}, 27, 2014.

\bibitem{he2016deep}
K.~He, X.~Zhang, S.~Ren, and J.~Sun.
\newblock Deep residual learning for image recognition.
\newblock In {\em Proceedings of the IEEE conference on computer vision and
  pattern recognition}, pages 770--778, 2016.

\bibitem{hesynthetic}
R.~He, S.~Sun, X.~Yu, C.~Xue, W.~Zhang, P.~Torr, S.~Bai, and X.~QI.
\newblock Is synthetic data from generative models ready for image recognition?
\newblock In {\em The Eleventh International Conference on Learning
  Representations}.

\bibitem{ho2020denoising}
J.~Ho, A.~Jain, and P.~Abbeel.
\newblock Denoising diffusion probabilistic models.
\newblock {\em Advances in neural information processing systems},
  33:6840--6851, 2020.

\bibitem{hou2007saliency}
X.~Hou and L.~Zhang.
\newblock Saliency detection: A spectral residual approach.
\newblock In {\em 2007 IEEE Conference on computer vision and pattern
  recognition}, pages 1--8. Ieee, 2007.

\bibitem{hu2022lora}
E.~J. Hu, Y.~Shen, P.~Wallis, Z.~Allen-Zhu, Y.~Li, S.~Wang, L.~Wang, W.~Chen,
  et~al.
\newblock Lora: Low-rank adaptation of large language models.
\newblock {\em ICLR}, 1(2):3, 2022.

\bibitem{huang2021snapmix}
S.~Huang, X.~Wang, and D.~Tao.
\newblock Snapmix: Semantically proportional mixing for augmenting fine-grained
  data.
\newblock In {\em Proceedings of the AAAI Conference on Artificial
  Intelligence}, volume~35, pages 1628--1636, 2021.

\bibitem{islam2024diffusemix}
K.~Islam, M.~Z. Zaheer, A.~Mahmood, and K.~Nandakumar.
\newblock Diffusemix: Label-preserving data augmentation with diffusion models.
\newblock In {\em Proceedings of the IEEE/CVF Conference on Computer Vision and
  Pattern Recognition}, pages 27621--27630, 2024.

\bibitem{itti2002model}
L.~Itti, C.~Koch, and E.~Niebur.
\newblock A model of saliency-based visual attention for rapid scene analysis.
\newblock {\em IEEE Transactions on pattern analysis and machine intelligence},
  20(11):1254--1259, 2002.

\bibitem{kang2023guidedmixup}
M.~Kang and S.~Kim.
\newblock Guidedmixup: an efficient mixup strategy guided by saliency maps.
\newblock In {\em Proceedings of the AAAI conference on artificial
  intelligence}, volume~37, pages 1096--1104, 2023.

\bibitem{khosla2011novel}
A.~Khosla, N.~Jayadevaprakash, B.~Yao, and F.-F. Li.
\newblock Novel dataset for fine-grained image categorization: Stanford dogs.
\newblock In {\em Proc. CVPR workshop on fine-grained visual categorization
  (FGVC)}, volume~2, 2011.

\bibitem{kim2020puzzle}
J.-H. Kim, W.~Choo, and H.~O. Song.
\newblock Puzzle mix: Exploiting saliency and local statistics for optimal
  mixup.
\newblock In {\em International conference on machine learning}, pages
  5275--5285. PMLR, 2020.

\bibitem{kingma2013auto}
D.~P. Kingma, M.~Welling, et~al.
\newblock Auto-encoding variational bayes.

\bibitem{krause20133d}
J.~Krause, M.~Stark, J.~Deng, and L.~Fei-Fei.
\newblock 3d object representations for fine-grained categorization.
\newblock In {\em Proceedings of the IEEE international conference on computer
  vision workshops}, pages 554--561, 2013.

\bibitem{kumari2023multi}
N.~Kumari, B.~Zhang, R.~Zhang, E.~Shechtman, and J.-Y. Zhu.
\newblock Multi-concept customization of text-to-image diffusion.
\newblock In {\em Proceedings of the IEEE/CVF conference on computer vision and
  pattern recognition}, pages 1931--1941, 2023.

\bibitem{litjens2017survey}
G.~Litjens, T.~Kooi, B.~E. Bejnordi, A.~A.~A. Setio, F.~Ciompi, M.~Ghafoorian,
  J.~A. Van Der~Laak, B.~Van~Ginneken, and C.~I. S{\'a}nchez.
\newblock A survey on deep learning in medical image analysis.
\newblock {\em Medical image analysis}, 42:60--88, 2017.

\bibitem{liu2019large}
Z.~Liu, Z.~Miao, X.~Zhan, J.~Wang, B.~Gong, and S.~X. Yu.
\newblock Large-scale long-tailed recognition in an open world.
\newblock In {\em Proceedings of the IEEE/CVF conference on computer vision and
  pattern recognition}, pages 2537--2546, 2019.

\bibitem{maji2013fine}
S.~Maji, E.~Rahtu, J.~Kannala, M.~Blaschko, and A.~Vedaldi.
\newblock Fine-grained visual classification of aircraft.
\newblock 2013.

\bibitem{mengsdedit}
C.~Meng, Y.~He, Y.~Song, J.~Song, J.~Wu, J.-Y. Zhu, and S.~Ermon.
\newblock Sdedit: Guided image synthesis and editing with stochastic
  differential equations.
\newblock In {\em International Conference on Learning Representations}.

\bibitem{muller2019does}
R.~M{\"u}ller, S.~Kornblith, and G.~E. Hinton.
\newblock When does label smoothing help?
\newblock {\em Advances in neural information processing systems}, 32, 2019.

\bibitem{mumuni2022data}
A.~Mumuni and F.~Mumuni.
\newblock Data augmentation: A comprehensive survey of modern approaches.
\newblock {\em Array}, 16:100258, 2022.

\bibitem{nichol2021improved}
A.~Q. Nichol and P.~Dhariwal.
\newblock Improved denoising diffusion probabilistic models.
\newblock In {\em International conference on machine learning}, pages
  8162--8171. PMLR, 2021.

\bibitem{nichol2022glide}
A.~Q. Nichol, P.~Dhariwal, A.~Ramesh, P.~Shyam, P.~Mishkin, B.~Mcgrew,
  I.~Sutskever, and M.~Chen.
\newblock Glide: Towards photorealistic image generation and editing with
  text-guided diffusion models.
\newblock In {\em International Conference on Machine Learning}, pages
  16784--16804. PMLR, 2022.

\bibitem{nilsback2008automated}
M.-E. Nilsback and A.~Zisserman.
\newblock Automated flower classification over a large number of classes.
\newblock In {\em 2008 Sixth Indian conference on computer vision, graphics \&
  image processing}, pages 722--729. IEEE, 2008.

\bibitem{otsu1975threshold}
N.~Otsu et~al.
\newblock A threshold selection method from gray-level histograms.
\newblock {\em Automatica}, 11(285-296):23--27, 1975.

\bibitem{park2022majority}
S.~Park, Y.~Hong, B.~Heo, S.~Yun, and J.~Y. Choi.
\newblock The majority can help the minority: Context-rich minority
  oversampling for long-tailed classification.
\newblock In {\em Proceedings of the IEEE/CVF conference on computer vision and
  pattern recognition}, pages 6887--6896, 2022.

\bibitem{patel2024conceptbed}
M.~Patel, T.~Gokhale, C.~Baral, and Y.~Yang.
\newblock Conceptbed: Evaluating concept learning abilities of text-to-image
  diffusion models.
\newblock In {\em Proceedings of the AAAI Conference on Artificial
  Intelligence}, volume~38, pages 14554--14562, 2024.

\bibitem{qi2024deadiff}
T.~Qi, S.~Fang, Y.~Wu, H.~Xie, J.~Liu, L.~Chen, Q.~He, and Y.~Zhang.
\newblock Deadiff: An efficient stylization diffusion model with disentangled
  representations.
\newblock In {\em Proceedings of the IEEE/CVF Conference on Computer Vision and
  Pattern Recognition}, pages 8693--8702, 2024.

\bibitem{qin2024sumix}
H.~Qin, X.~Jin, H.~Zhu, H.~Liao, M.~A. El-Yacoubi, and X.~Gao.
\newblock Sumix: Mixup with semantic and uncertain information.
\newblock In {\em European Conference on Computer Vision}, pages 70--88.
  Springer, 2024.

\bibitem{qin2020resizemix}
J.~Qin, J.~Fang, Q.~Zhang, W.~Liu, X.~Wang, and X.~Wang.
\newblock Resizemix: Mixing data with preserved object information and true
  labels.
\newblock {\em arXiv preprint arXiv:2012.11101}, 2020.

\bibitem{qiu2023controlling}
Z.~Qiu, W.~Liu, H.~Feng, Y.~Xue, Y.~Feng, Z.~Liu, D.~Zhang, A.~Weller, and
  B.~Sch{\"o}lkopf.
\newblock Controlling text-to-image diffusion by orthogonal finetuning.
\newblock {\em Advances in Neural Information Processing Systems},
  36:79320--79362, 2023.

\bibitem{radford2021learning}
A.~Radford, J.~W. Kim, C.~Hallacy, A.~Ramesh, G.~Goh, S.~Agarwal, G.~Sastry,
  A.~Askell, P.~Mishkin, J.~Clark, et~al.
\newblock Learning transferable visual models from natural language
  supervision.
\newblock In {\em International conference on machine learning}, pages
  8748--8763. PmLR, 2021.

\bibitem{ridnik1imagenet}
T.~Ridnik, E.~Ben-Baruch, A.~Noy, and L.~Zelnik-Manor.
\newblock Imagenet-21k pretraining for the masses.
\newblock In {\em Thirty-fifth Conference on Neural Information Processing
  Systems Datasets and Benchmarks Track (Round 1)}.

\bibitem{rombach2022high}
R.~Rombach, A.~Blattmann, D.~Lorenz, P.~Esser, and B.~Ommer.
\newblock High-resolution image synthesis with latent diffusion models.
\newblock In {\em Proceedings of the IEEE/CVF conference on computer vision and
  pattern recognition}, pages 10684--10695, 2022.

\bibitem{ronneberger2015u}
O.~Ronneberger, P.~Fischer, and T.~Brox.
\newblock U-net: Convolutional networks for biomedical image segmentation.
\newblock In {\em Medical image computing and computer-assisted
  intervention--MICCAI 2015: 18th international conference, Munich, Germany,
  October 5-9, 2015, proceedings, part III 18}, pages 234--241. Springer, 2015.

\bibitem{rother2004grabcut}
C.~Rother, V.~Kolmogorov, and A.~Blake.
\newblock " grabcut" interactive foreground extraction using iterated graph
  cuts.
\newblock {\em ACM transactions on graphics (TOG)}, 23(3):309--314, 2004.

\bibitem{ruiz2023dreambooth}
N.~Ruiz, Y.~Li, V.~Jampani, Y.~Pritch, M.~Rubinstein, and K.~Aberman.
\newblock Dreambooth: Fine tuning text-to-image diffusion models for
  subject-driven generation.
\newblock In {\em Proceedings of the IEEE/CVF conference on computer vision and
  pattern recognition}, pages 22500--22510, 2023.

\bibitem{russakovsky2015imagenet}
O.~Russakovsky, J.~Deng, H.~Su, J.~Krause, S.~Satheesh, S.~Ma, Z.~Huang,
  A.~Karpathy, A.~Khosla, M.~Bernstein, et~al.
\newblock Imagenet large scale visual recognition challenge.
\newblock {\em International journal of computer vision}, 115:211--252, 2015.

\bibitem{sagawadistributionally}
S.~Sagawa, P.~W. Koh, T.~B. Hashimoto, and P.~Liang.
\newblock Distributionally robust neural networks.
\newblock In {\em International Conference on Learning Representations}.

\bibitem{saharia2022photorealistic}
C.~Saharia, W.~Chan, S.~Saxena, L.~Li, J.~Whang, E.~L. Denton, K.~Ghasemipour,
  R.~Gontijo~Lopes, B.~Karagol~Ayan, T.~Salimans, et~al.
\newblock Photorealistic text-to-image diffusion models with deep language
  understanding.
\newblock {\em Advances in neural information processing systems},
  35:36479--36494, 2022.

\bibitem{samuel2021generalized}
D.~Samuel, Y.~Atzmon, and G.~Chechik.
\newblock From generalized zero-shot learning to long-tail with class
  descriptors.
\newblock In {\em Proceedings of the IEEE/CVF winter conference on applications
  of computer vision}, pages 286--295, 2021.

\bibitem{selvaraju2017grad}
R.~R. Selvaraju, M.~Cogswell, A.~Das, R.~Vedantam, D.~Parikh, and D.~Batra.
\newblock Grad-cam: Visual explanations from deep networks via gradient-based
  localization.
\newblock In {\em Proceedings of the IEEE international conference on computer
  vision}, pages 618--626, 2017.

\bibitem{sengupta2020background}
S.~Sengupta, V.~Jayaram, B.~Curless, S.~M. Seitz, and
  I.~Kemelmacher-Shlizerman.
\newblock Background matting: The world is your green screen.
\newblock In {\em Proceedings of the IEEE/CVF Conference on Computer Vision and
  Pattern Recognition}, pages 2291--2300, 2020.

\bibitem{shorten2019survey}
C.~Shorten and T.~M. Khoshgoftaar.
\newblock A survey on image data augmentation for deep learning.
\newblock {\em Journal of big data}, 6(1):1--48, 2019.

\bibitem{simonyan2014deep}
K.~Simonyan, A.~Vedaldi, and A.~Zisserman.
\newblock Deep inside convolutional networks: visualising image classification
  models and saliency maps.
\newblock International Conference on Learning Representations, 2014.

\bibitem{songdenoising}
J.~Song, C.~Meng, and S.~Ermon.
\newblock Denoising diffusion implicit models.
\newblock In {\em International Conference on Learning Representations}.

\bibitem{songscore}
Y.~Song, J.~Sohl-Dickstein, D.~P. Kingma, A.~Kumar, S.~Ermon, and B.~Poole.
\newblock Score-based generative modeling through stochastic differential
  equations.
\newblock In {\em International Conference on Learning Representations}.

\bibitem{trabuccoeffective}
B.~Trabucco, K.~Doherty, M.~Gurinas, and R.~Salakhutdinov.
\newblock Effective data augmentation with diffusion models.
\newblock In {\em R0-FoMo: Robustness of Few-shot and Zero-shot Learning in
  Large Foundation Models}.

\bibitem{tumanyan2023plug}
N.~Tumanyan, M.~Geyer, S.~Bagon, and T.~Dekel.
\newblock Plug-and-play diffusion features for text-driven image-to-image
  translation.
\newblock In {\em Proceedings of the IEEE/CVF Conference on Computer Vision and
  Pattern Recognition}, pages 1921--1930, 2023.

\bibitem{uddinsaliencymix}
A.~S. Uddin, M.~S. Monira, W.~Shin, T.~Chung, and S.-H. Bae.
\newblock Saliencymix: A saliency guided data augmentation strategy for better
  regularization.
\newblock In {\em International Conference on Learning Representations}.

\bibitem{wah2011caltech}
C.~Wah, S.~Branson, P.~Welinder, P.~Perona, and S.~Belongie.
\newblock The caltech-ucsd birds-200-2011 dataset.
\newblock 2011.

\bibitem{wang2024enhance}
Z.~Wang, L.~Wei, T.~Wang, H.~Chen, Y.~Hao, X.~Wang, X.~He, and Q.~Tian.
\newblock Enhance image classification via inter-class image mixup with
  diffusion model.
\newblock In {\em Proceedings of the IEEE/CVF Conference on Computer Vision and
  Pattern Recognition}, pages 17223--17233, 2024.

\bibitem{ye2023exploiting}
M.~Ye-Bin, N.~Hyeon-Woo, W.~Choi, N.~Kim, S.~Kwak, and T.-H. Oh.
\newblock Exploiting synthetic data for data imbalance problems: baselines from
  a data perspective.
\newblock {\em arXiv preprint arXiv:2308.00994}, 6, 2023.

\bibitem{yu2018generative}
J.~Yu, Z.~Lin, J.~Yang, X.~Shen, X.~Lu, and T.~S. Huang.
\newblock Generative image inpainting with contextual attention.
\newblock In {\em Proceedings of the IEEE conference on computer vision and
  pattern recognition}, pages 5505--5514, 2018.

\bibitem{yun2019cutmix}
S.~Yun, D.~Han, S.~J. Oh, S.~Chun, J.~Choe, and Y.~Yoo.
\newblock Cutmix: Regularization strategy to train strong classifiers with
  localizable features.
\newblock In {\em Proceedings of the IEEE/CVF international conference on
  computer vision}, pages 6023--6032, 2019.

\bibitem{zhang2018mixup}
H.~Zhang, M.~Cisse, Y.~N. Dauphin, and D.~Lopez-Paz.
\newblock mixup: Beyond empirical risk minimization.
\newblock In {\em International Conference on Learning Representations}, 2018.

\bibitem{zhao2015saliency}
R.~Zhao, W.~Ouyang, H.~Li, and X.~Wang.
\newblock Saliency detection by multi-context deep learning.
\newblock In {\em Proceedings of the IEEE conference on computer vision and
  pattern recognition}, pages 1265--1274, 2015.

\bibitem{zhou2016learning}
B.~Zhou, A.~Khosla, A.~Lapedriza, A.~Oliva, and A.~Torralba.
\newblock Learning deep features for discriminative localization.
\newblock In {\em Proceedings of the IEEE conference on computer vision and
  pattern recognition}, pages 2921--2929, 2016.

\bibitem{zhou2017places}
B.~Zhou, A.~Lapedriza, A.~Khosla, A.~Oliva, and A.~Torralba.
\newblock Places: A 10 million image database for scene recognition.
\newblock {\em IEEE transactions on pattern analysis and machine intelligence},
  40(6):1452--1464, 2017.

\end{thebibliography}
}

\end{document}